\documentclass[review]{elsarticle}

\usepackage{bm}
\usepackage{multirow}
\usepackage{amsmath}
\usepackage{amssymb}
\usepackage{float}
\usepackage{booktabs}
\usepackage{CJK}
\usepackage{array,color}
\usepackage{graphicx}
\usepackage{xcolor}
\usepackage{subfigure}
\usepackage[pagebackref=false,breaklinks=false,letterpaper=false,colorlinks,bookmarks=false]{hyperref}

\journal{Journal of \LaTeX\ Templates}

\begin{document}

\begin{frontmatter}

\title{Hybrid-Attention Guided Network with Multiple Resolution Features for Person Re-Identification}


\author[mymainaddress]{Guoqing Zhang}

\author[mymainaddress]{Junchuan Yang}
\author[mymainaddress]{Yuhui Zheng}
\author[mysecondaryaddress]{Ye Wang\corref{mycorrespondingauthor}}
\cortext[mycorrespondingauthor]{Corresponding author}
\ead{wangye@nufe.edu.cn}
\author[mythirdaryaddress]{Yi Wu}
\author[myfourtharyaddress]{Shengyong Chen}

\address[mymainaddress]{School of Computer and Software, Nanjing University of Information Science and Technology, Nanjing, 210044, China}
\address[mysecondaryaddress]{School of Marketing and Logistics Management, Nanjing University of Finance and Economics, Nanjing, China}
\address[mythirdaryaddress]{Wormpex AI Research}
\address[myfourtharyaddress]{School of Computer Science and Engineering, Tianjin University of Technology}

\begin{abstract}
Extracting effective and discriminative features is very important for addressing the challenges of person re-identification (re-ID). Prevailing deep convolutional neural networks usually use high-level features for identifying pedestrian. However, some essential spatial information resided in low-level features will be lost when learning the high-level features. Most existing person re-ID methods mainly rely on hand-craft bounding boxes, where person images are precisely aligned. It is unrealistic in practical applications due to the inaccuracy of automatic detection algorithms. To address these problems, we propose a hybrid-attention guided network with multiple resolution features for person re-ID. We first construct a multi-resolution fusion strategy to ensure that multi-resolution features can be spatially aligned during feature fusion and at the same time ensure the discriminability of features after fusion. Then, we introduce the attention mechanism and multi-granularity operation to reduce the impact of irregular bounding boxes by gently expanding the size of feature maps. In addition, a new multi-pool feature extractor is designed to obtain different types of information by using two different pools and the feature representation capability can be further improved. Extensive experiments display the superiority of our approach. Our code is available at \url{https://github.com/libraflower/MutipleFeature-for-PRID}.
\end{abstract}


\end{frontmatter}


\section{Introduction}
Given a query image of a person, re-identification is to identify the person from different cameras without overlapping fields of view. Due to complicated and variable conditions such as person posture, camera view, and low resolution, person re-ID still encounters great challenges. Currently, existing person re-ID approaches mainly contain two aspects: traditional algorithms and deep learning-based algorithms. Most traditional person re-ID approaches are mainly based on hand-craft features, which cannot adapt well to complex scenarios with a large amount of data. Recently, numerous deep learning based algorithms have been presented and greatly improved the performance of re-ID. Different from traditional methods, deep learning approaches can extract highly discriminative image features and learn more accurate similarity measurements by building multi-layer models with nonlinear characteristics.

\begin{figure}[!t]
 \centering
 \includegraphics[width=85mm]{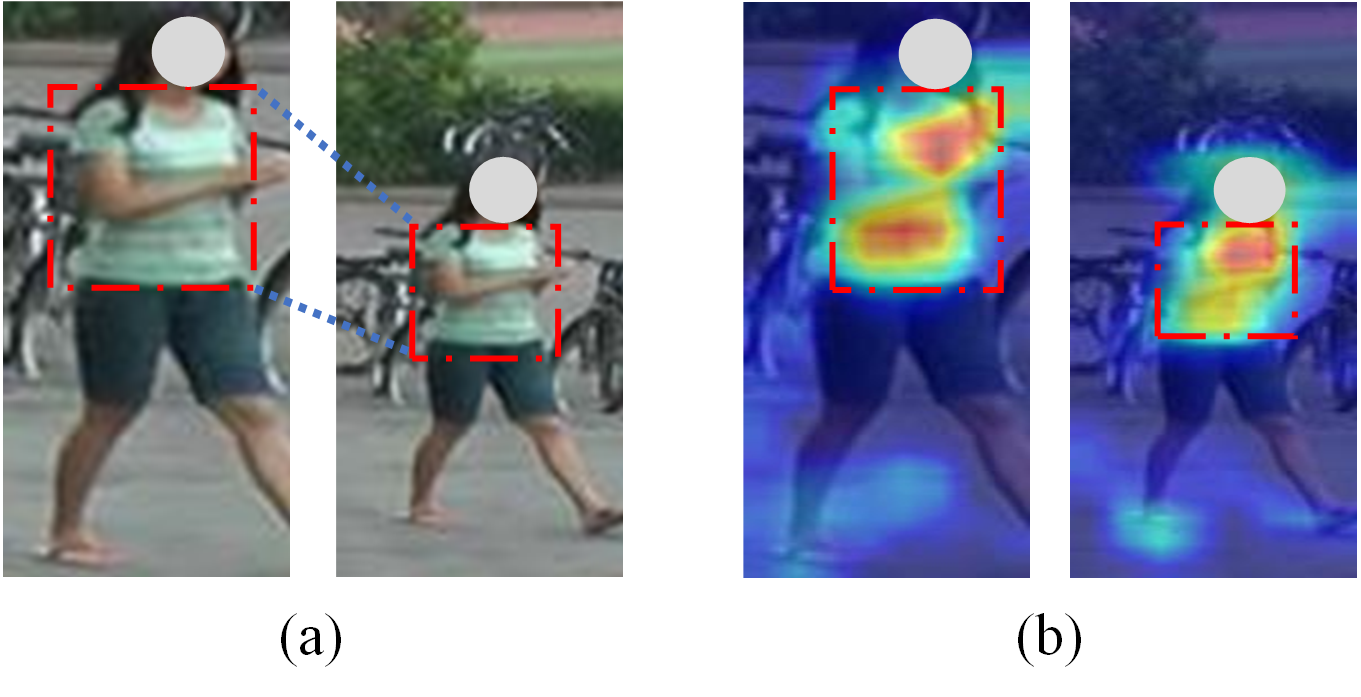}\\
 \caption{(a) The two images on the left represent two different images of the same person whose bounding boxes do not match perfectly. (b) The two images on the right show the activation map of our method. It can be seen that our method can still find more discriminative local regions (inside the red border) completely.}
  \label{fig:1}
\end{figure}

Although deep learning based person re-ID algorithms have received increasing explorations~\cite{8,21,34,39,56}, it is still very difficult to effectively apply them to real scenes. Prevailing algorithms are usually designed based on the manually annotated datasets, which provide accurate pedestrian detection bounding boxes. However, in practice, the automatically detected bounding boxes are usually blurred and misaligned, which seriously degrade the performances of existing re-ID algorithms (as shown in Figure~\ref{fig:1} (a)). Besides, due to some limitations of the deep models, some discriminative features will be lost with the deepening of the network, which brings difficulty to practical applications.

To match the person identity, features extracted from the top (deep) layer of a Convolutional Neural Network (CNN) are generally used for similarity matching. However, \textcolor{black}{one CNN} usually contains multiple feature extraction layers, which are superimposed layer by layer. From the shallow to the deep layer, the visual cues captured by high-level feature maps tend to be more abstract and the semantic levels are higher, i.e., features of the last layer of a network mostly encode semantic information, like object presence~\cite{29}, may lose some basic information, such as the color of clothes or the shape of human’s body. To remedy this problem, we should reason simultaneously across multiple levels of semantic abstraction. That means both deep features and shallow features of a CNN should be considered jointly.

Part-based models have shown superior performance in many computer vision tasks because of their robustness to image occlusion or partial variation challenges, which also affect the development of person re-ID. Part-based Convolutional Baseline (PCB) structure~\cite{39} is a simple and effective model, which is one of the most acknowledged part-based approaches, and its performance exceeds most deep learning models. However, the current PCB model has some shortcomings: (1) PCB structure only considers the divided features and completely ignores the role of global features in the whole network; (2) The effectiveness of PCB structure depends on the
accuracy of the bounding box and ignores the incomplete image detection and blurred bounding boxes in real scenes. In proposed method, we exploit multi-granularity operation to reduce the dependence of the model on the precise bounding boxes.

Recently, many researchers have introduced attention mechanisms into deep models to strengthen the learned features and suppress some disturbances of the features~\cite{17,54,60,61}.
The purpose of attention is to make the model pay more attention to the salient regions in the image during training process and ignore the noises that are not related to the recognition.
This is consistent with the task of person re-ID.
Therefore, attention mechanisms can be used to mine the significant regions of pedestrians to enhance the representation ability of the model.


To address these concerns, we propose a hybrid-attention guided network with multiple resolution features for person re-ID.
Firstly, we construct a multi-resolution feature fusion strategy, which aims to fuse high- and low-level features for alleviating the influence of information loss caused in the process of learning high-level features. Besides, we utilize hybrid-attention mechanism to suppress the noises and strength the salient regions in the features simultaneously.
Secondly, our proposed model adopts a multi-granularity operation so as to gradually expand the size of feature maps and reduce the dependence of model on the precise bounding boxes.
Finally, a multi-pool feature extractor is designed to further enhance the discriminant ability of features by fusing two types of pooling information.

Our main contributions are summarized as follows:


\begin{itemize}

\item  We propose a hybrid-attention guided network with multiple resolution features for person re-ID, which constructs a multi-resolution feature fusion strategy to ensure the spatial alignment of feature maps at different stages during feature fusion. In addition, this strategy can also explore the potential salient regions in the feature maps through the hybrid-attention mechanism and can be suitable for any convolutional neural network.
\item  We design a new multi-pool feature extractor to further improve the feature representation capability of the model by using two different pooling modes to obtain multiple discriminant information existing in the feature maps.
\item  Our model achieves the state-of-the-art performance on four benchmark datasets, including Market-1501, DukeMTMC-reID, CUHK03-NP and MSMT17.

\end{itemize}

\section{Related Work}

\subsection{Deep Learning for Person Re-ID}
Person re-ID is usually regarded as a sub-problem of image retrieval~\cite{18}. It refers to the task of finding the target in different periods of videos captured from different cameras in different locations under the condition of a given pedestrian. Traditional person re-ID approaches mainly focus on learning better manual visual features to reduce intra-person divergence and enhance the inter-person discrimination capability~\cite{1,4,6} or designing appropriate distance metric to accurately measure the similarity of different identities~\cite{2,3,5,7,72}. In recent years, with the extensive progress of deep learning technology, it has been widely adopted to computer vision tasks, such as medical image processing and person re-ID~\cite{9,10,12,11,13,16}. Ding et al.~\cite{new2,new1} developed deep learning to the classification of different longitudinal surfaces of infant cerebrum fMRI and multimodal infant brain segmentation problems. Besides, different from traditional re-ID approaches, methods based on deep learning can adaptively learn discriminative features from pedestrian images~\cite{11,17,46} and contribute a lot to the re-ID task.

Previous person re-ID methods use global features of the whole image to match images~\cite{21,22,25,43}, which ignore various partial information of a given person. Recently, researchers found that local features show a great prospect for dealing with the misalignment of the pedestrian images. Therefore, extracting various local features from a large-scale dataset to improve re-ID performance is now the new mainstream ~\cite{14,16,17,24,10,11}. The idea of extracting different solutions at multiple stages and using skip connections to combine \textcolor{black}{multiple-stage} features has proven to be very useful in the field of target detection~\cite{31,33}, image classification, and object recognition. This idea has also been exploited in person re-ID and promoted re-ID model to obtain very accurate prediction \textcolor{black}{results}~\cite{40,41}. However, previous person re-ID methods usually pool the features of different stages into a single vector, which will inevitably lose some feature information of different stages~\cite{34}. Therefore to reduce this loss, we keep the size of feature maps from previous stages the same as that of the last stage.

\subsection{Local Features for Person Re-ID}
Global and local features extracted by CNNs are widely used to perform person matching. The global \textcolor{black}{feature} means the network extracts \textcolor{black}{information} from the whole image without considering all local information~\cite{47,34}. Due to the complexity of pedestrian images, a single global feature cannot meet the performance requirement, so learning more complex local features has become a research hotspot~\cite{16,17,24,28,29}.

Feature map partition is a very common feature extraction algorithm~\cite{24,39,15,13}. Most of the current methods adopt integrated global features and many stripe-based features to achieve advanced performance by linear pooling the partitioned feature maps~\cite{16,26,30,71}. In~\cite{68}, a multi-level network was proposed to decompose human features into multiple fine-grained local features, and fine-grained local features of different levels were integrated at the end of the model. Tay et al.~\cite{17} used three independent structures to obtain local features, global features, and pose features respectively, and then fused these three kinds of information to obtain \textcolor{black}{the} final feature.

\subsection{Attention Mechanism for Person Re-ID}
Attention mechanism has been widely used in a variety of deep learning domains, such as image processing, natural language processing, speech recognition, and so on. The human visual system tends to pay attention to some information of assist judgment in the image and ignore the irrelevant information. Therefore, in computer vision community, some parts of the input may be more useful to the target task than others. The attention mechanism allows the model to adaptively focus on certain parts of the input related to the target.

Attention mechanism has also been adopted to address the re-ID task~\cite{17,61,60,63,49,39,54}. Many models hope to find out more discriminative features (color or texture) in the image through attention \textcolor{black}{mechanism} and ignore other features (background) which are irrelevant to \textcolor{black}{the} task. Sun et al.~\cite{27} simply used the attention mechanism to locate the visible areas in a given image through a positioning method and learned their local features based on these visible areas. \textcolor{black}{Chen et al.~\cite{49} proposed the high-order attention module and \textcolor{black}{utilized} the complex high-order statistical information in the attention mechanism to capture the subtle differences between pedestrians. Huang et al.~\cite{54} introduced 3D attention into the model to enhance the ability of feature discrimination.}

\section{The Proposed Method}

Previous person Re-ID methods usually adopted the highest-level features for representing \textcolor{black}{pedestrians}, such as the output of the last convolutional layer of \textcolor{black}{ResNet-50}. Although it is very useful in representing object images using high-level features, yet at the same time some low-level features such as texture and color information will be lost. These low-level features are very significant clues for performing person re-ID. More importantly, in the last several layers of \textcolor{black}{CNN}, the distinguishing rate of feature \textcolor{black}{maps} is lower, which may not accurately represent the patterns on clothes, facial subtle features, differences in posture \textcolor{black}{or} other details. This indicates that person re-ID methods benefit from information fusion with multiple layers.




\subsection{Multi-resolution Feature Fusion Strategy}
\textcolor{black}{A Multi-resolution Feature Fusion strategy (MFF) can fuse different resolution features of multiple stages in any CNN. For a convolutional network, the receptive field of deep feature map is usually larger, while the receptive field of low feature map is usually smaller. Therefore, the feature maps with different resolutions are not aligned in spatial. \textcolor{black}{It is necessary to align these feature maps in spatial} before fusion to ensure that the feature representation ability of the model is not affected. Since pooling has the characteristics of translation invariance and max-pooling can preserve \textcolor{black}{the most responsive and strongest part of the information in feature maps, we use max-pooling to ensure that feature maps} of different resolutions are spatially aligned before fusion.}

\begin{figure*}[!t]
\centering
\includegraphics[width=122mm]{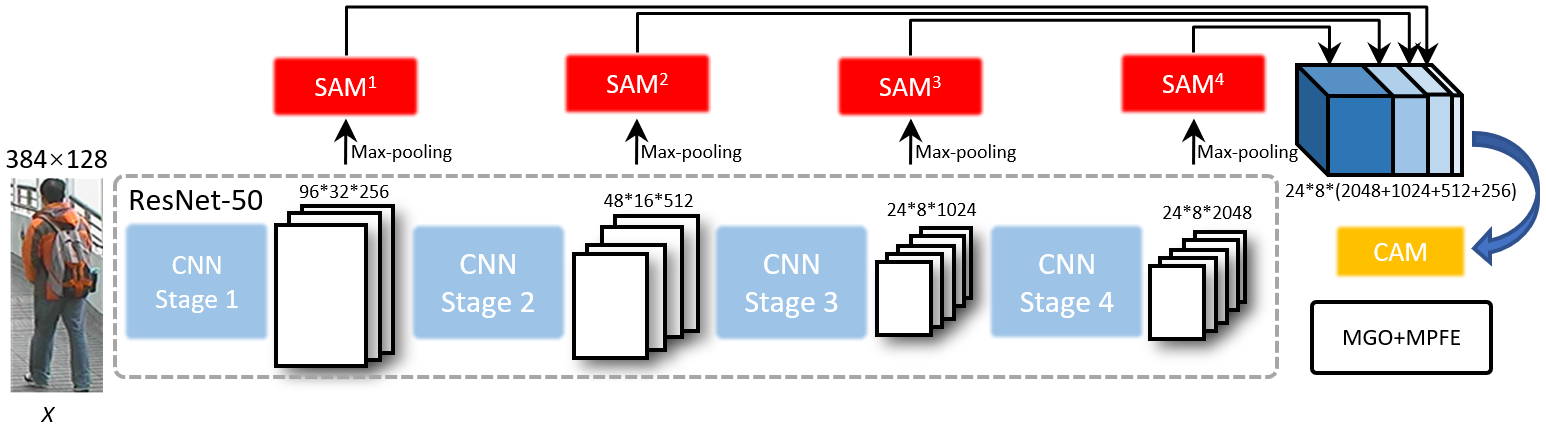}\\
\caption{Architecture of our model for person re-ID. Our proposed model consists of ResNet-50, SAM, CAM, MGO and MPFE. We extracted features from different stages of the network (ResNet-50), weighted them through the spatial attention module (SAM), spliced them together by channels, and then weighted the spliced feature maps through the channel attention module (CAM), followed by multi-granularity feature extraction (MGO+MPFE).}
  \label{fig:2}
\end{figure*}

\textcolor{black}{As shown in Figure~\ref{fig:2}, in order to better integrate the high and low-level features of the CNNs, we adopt the hybrid-attention mechanism to enhance the discriminative regions and suppress the noises in these features. More specifically, we use spatial attention module (SAM) to mine the discriminative regions of the spatial-aligned feature maps in spatial dimension before the fusion. The SAM is mainly concerned with "where" is the feature that is helpful for recognition. Since the cascade during the fusion expands the channel dimension of feature maps, we use channel attention module (CAM) to find out "which" channels we need after the fusion.}

\subsection{Hybrid-Attention Mechanism}

Hybrid-Attention Mechanism (HAM) \textcolor{black}{consists of a CAM and a SAM, in which the CAM explores the correlation between channels and the SAM explores the strong semantic features within the spatial dimension.}

\begin{figure}[]
 \centering
 \includegraphics[width=90mm]{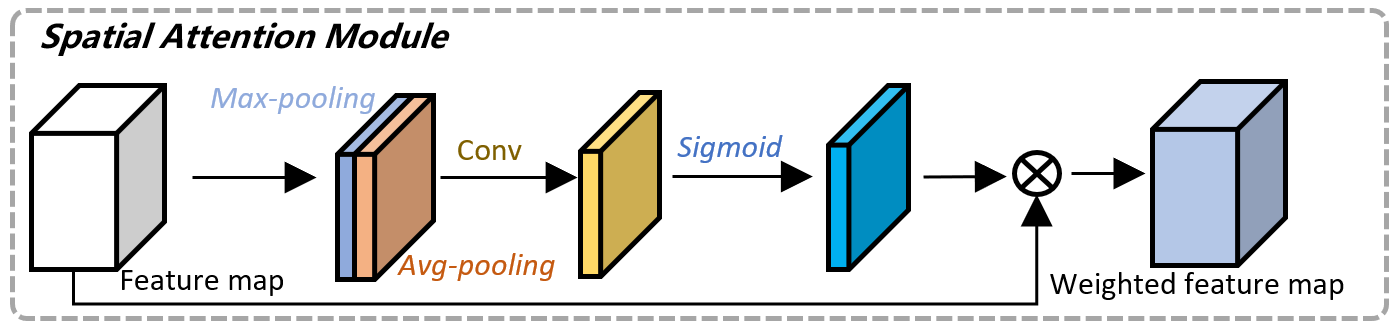}\\
 \caption{Pipeline of Spatial Attention Module (SAM).}
  \label{fig:SAM}
\end{figure}

\textbf{Spatial Attention Module (SAM)}. We exploit the relationship of feature space to generate spatial attention map.
The pipeline of SAM is shown in Figure~\ref{fig:SAM}. By using two different pooling operations, two two-dimensional \textcolor{black}{context descriptors} from the original \textcolor{black}{feature maps} can be generated, denoted as $F_{max}^{SAM}$ and $F_{avg}^{SAM}$. Then we connect these two features and generate the final two-dimensional spatial attention map by using the convolution \textcolor{black}{layer} and sigmoid function. Thus, the finally \textcolor{black}{spatial} attention can be defined as:
\begin{equation}
\begin{aligned}
M_{SAM}(F)&=\sigma(f([AvgPool(F)\textcolor{black}{\leftarrow} MaxPool(F)]))\\
&=\sigma(f([F_{max}^{SAM}\textcolor{black}{\leftarrow} F_{avg}^{SAM}])),
\end{aligned}
\end{equation}where $\sigma$ is the sigmoid function, $f$ means the convolution layer and \textcolor{black}{$A \leftarrow B$ means to connect $B$ to the end of $A$ by channel.}

\begin{figure}[]
 \centering
 \includegraphics[width=90mm]{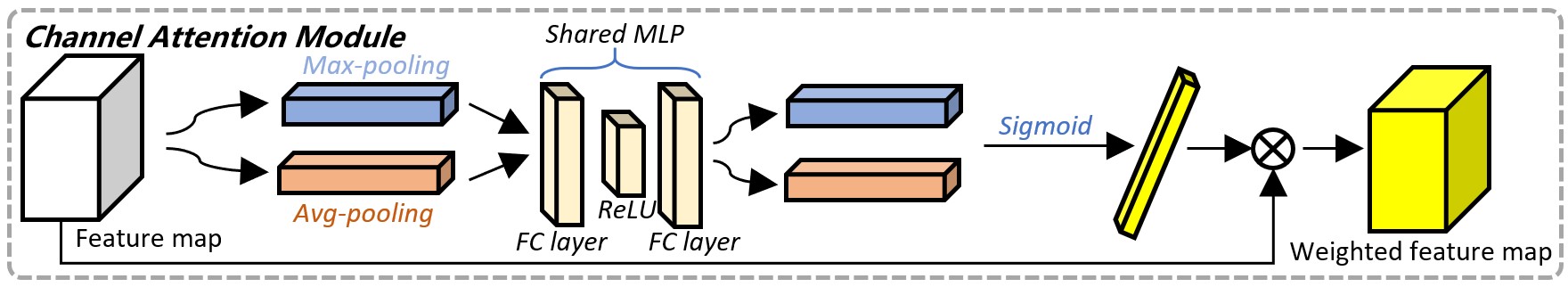}\\
 \caption{Pipeline of Channel Attention Module (CAM).}
  \label{fig:CAM}
\end{figure}

\textbf{Channel Attention Module (CAM)}. The channel attention map can be generated through the inter-channel \textcolor{black}{relationship} of features.
As shown in Figure~\ref{fig:CAM}, we first get different \textcolor{black}{1D context descriptors} through the operations of max-pooling and avg-pooling, denoted as $F_{max}^{CAM}$ and $F_{avg}^{CAM}$. And then we obtain the channel attention map through a shared multi-layer \textcolor{black}{perception}, which includes two \textcolor{black}{fully connected layers} and \textcolor{black}{ReLU} activation functions. The final \textcolor{black}{channel attention confidence map} can be briefly expressed as:
\begin{equation}
\begin{aligned}
M_{CAM}(F)&=\sigma(f_{multi}([AvgPool(F)+MaxPool(F)]))\\
&=\sigma(f_{multi}([F_{avg}^{CAM}+F_{max}^{CAM}])),
\end{aligned}
\end{equation}where $\sigma$ represents the sigmoid function and $f_{multi}$ represents the shared multi-layer \textcolor{black}{perception}.

\subsection{Multi-Granularity Operation}

Multi-Granularity Operation (MGO) module aims to obtain local features \textcolor{black}{at different levels of feature maps to alleviate the influence of inaccurate bounding boxes on the model.}

We first obtain feature $F\in R(C*H*W)$ \textcolor{black}{\textcolor{black}{through} the fusion and HAM (as shown in Figure~\ref{fig:2}).}
\textcolor{black}{Then, \textcolor{black}{we} set the partitioning level to $K$ and get local features of different granularity in different levels. At the top level of MGO, only one global feature is included. Similarly, \textcolor{black}{at} the bottom layer, the module will obtain $k$ stripe-based local features: $F_{1\sim k}\in R(C* (H/k) *W)$. \textcolor{black}{At} the middle layer, the module will reorganize the feature maps according to level $p$ to obtain the relaxed feature maps:}
\textcolor{black}{
\begin{equation}
M_{P}=F_{1\sim P}\in R(C,(((k-1)*H/n+1):((k-1)*H/n +l*H/n)),W),
\end{equation}
where $l=1,2...,p$ and $n= 1,2...,k$.}

\subsection{Multi-Pool Feature Extractor}
The local features obtained by the MGO will be sent to the Multi-Pool Feature Extractor (MPFE) \textcolor{black}{to obtain the final feature representation vector.} \textcolor{black}{For high-level features, avg-pooling can better aggregate the high-level semantic information of features.} However, max-pooling can \textcolor{black}{preserve the most responsive part of the feature maps.} Therefore, two different pooling methods are used in this extractor \textcolor{black}{to ensure the integrity of feature information and highlight the discriminative part of features.}

\begin{figure*}[!t]
  \centering
  \includegraphics[width=4.5in]{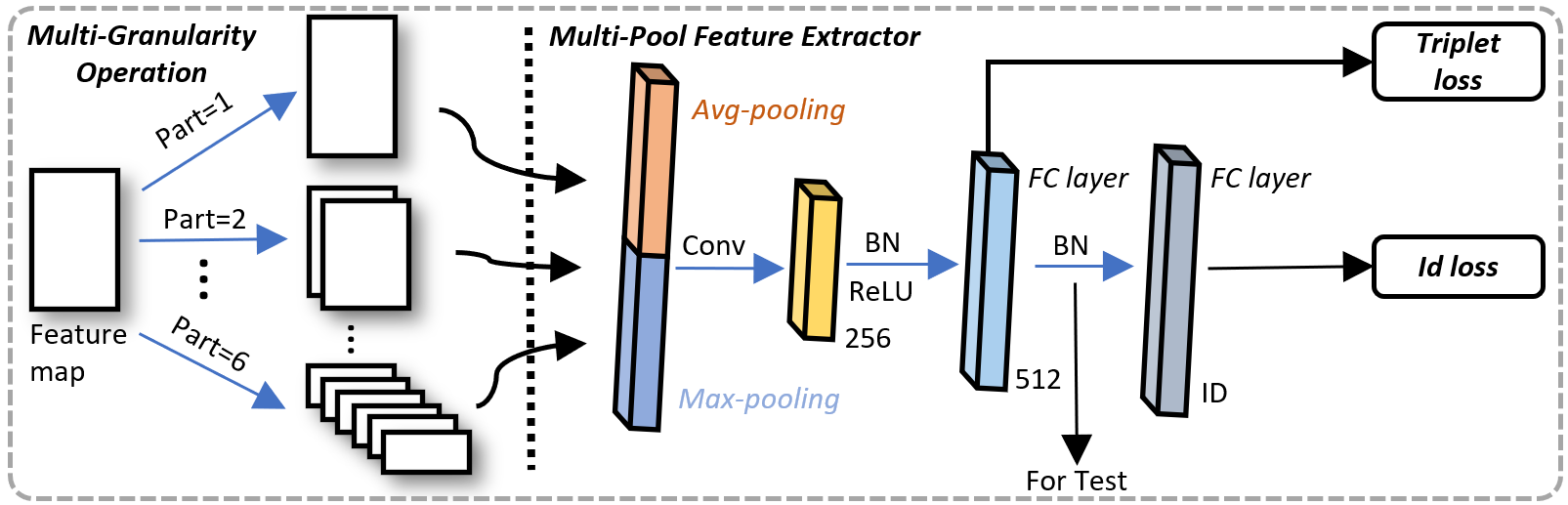}\\
  \caption{Illustration of multi-granularity operation (MGO) and multi-pool feature extractor (MPFE). The MGO module is responsible for partitioning embeddings to obtain different granularity of local features. The MPFE module is used to extract the final feature.}
  \label{fig:5}
\end{figure*}

As shown in Figure~\ref{fig:5}, we can get two \textcolor{black}{corresponding context descriptors} by adopting the \textcolor{black}{max-pooling} and \textcolor{black}{avg-pooling} for the obtained \textcolor{black}{local} feature maps, and then splicing the two \textcolor{black}{descriptors}. As the dimension increases after splicing, a $1*1$ convolutional layer is introduced to reduce the dimension of the \textcolor{black}{spliced descriptor} \textcolor{black}{for improving the efficiency of the extractor}. After a \textcolor{black}{Batch Normalization} (BN) layer and a  \textcolor{black}{ReLU} activation function, the dimension of descriptor is changed into 512-dimensional through a \textcolor{black}{fully connected (FC)} layer, and this feature vector is used to calculate the triplet loss. Then, the final classification feature \textcolor{black}{vector} can be obtained through a BN layer and a \textcolor{black}{FC} layer, which is used to calculate ID loss. \textcolor{black}{Note that, in the testing phase, we use the output of BN layer as the testing vector. }



\subsection{Multiple Loss Function}
In order to learn discriminant features, two different loss functions, including ID loss~\cite{4} and triplet loss~\cite{23} are exploited in our network. Therefore, we define the total loss of our network as:
\begin{equation}
\begin{aligned}
L^{a l l}=\frac{1}{N}L^{i d}+ \frac{1}{M}L^{t p},
\end{aligned}
\end{equation}where \textcolor{black}{$L^{i d}$ is ID loss, $L^{t p}$ is triplet loss.} $N$ represents the number of ID loss calculated, and $M$ represents the number of triplet loss.

\section{Experiments}
To verify the superiority of our model, four datasets such as Market-1501\cite{35}, DukeMTMC-reID\cite{19} , CUHK03\cite{36} and MSMT17\cite{70} are used in our experiments.

\subsection{Experimental Setting}
\textbf{Implementation details:} \textcolor{black}{Following} the experiment setting in~\cite{39}, all images of input are resized to $384*128$. In our model, we utilize the ResNet-50 network with the pretrained weights on ImageNet as the backbone \textcolor{black}{and remove} the last full connection layer. \textcolor{black}{To retain more spatial details, we change strides of last sampling to 1 in last stage}, such that the size of the feature map we obtained from the stage 4 of CNN is 2048*24*8.

In all experiments, we set the number of parts as 6 and the margin in the triplet loss is 1.0 in our network. Besides, label smoothing strategy is used in ID loss. We utilize the common data augmentation strategies to improve the performance of our model, including random erasing, horizontal flipping, random cropping and dropblock~\cite{62}. The size of mini-batch is 64 for each iteration. We use stochastic gradient descent (SGD) as the optimizer with a momentum of 0.9 and 0.0005 is the weight decay factor. We train our model for 150 epochs and set 0.01 as the initial learning rate, and later the warm-up strategy as described in equal (5) is adopted. We keep the same experiments setting on all datasets. \textcolor{blue}{With a NVIDIA Tesla A100 GPU and Pytorch as the platform, training ResNet-50 (IDE) and our model on Market-1501 (12,936 training images) consumes about 100 and 180 minutes, respectively. The increased training time is mainly caused by the cancellation of the last spatial down-sample operation in the stage 4, which enlarges the feature map by 4×.}

The learning rate $lr(e)$ in epoch $e$ is computed as:
\begin{equation}
lr(e)=
\left\{
\begin{aligned}
&  3 * 10^{-4} * \frac{e}{10}, e\leq 10 \\
& 0.01, 10<e\leq 60 \\
& 0.005, 60<e \leq 90 \\
&0.0025, 90<e \leq 120 \\
& 0.00125, 120<e \leq 150
\end{aligned}
\right.
\end{equation}

\textbf{Evaluation metrics:} To compare our method with existing advanced approaches, we use Cumulative Match Characteristics (CMC) and mean Average Precision (mAP) to measure their performances on all the datasets. Notably, we do not adopt the re-ranking strategy to improve the results in our experiments.

\subsection{Datasets}
\textbf{Market-1501:} This dataset includes 1,501 different identities of 32,668 images observed from six cameras with overlapping and one camera is low-resolution, five cameras are high-resolution. Following the same setting in PCB~\cite{39}, 751 IDs with 12,936 images are allocated for training and the rest 750 IDs with 19,732 gallery images and 3,368 query images \textcolor{black}{build} the testing set.

\textbf{DukeMTMC-reID:} This dataset consists of 1,404 identities, 2,228 queries, 17,661 gallery images, and 16,522 training images captured from 8 high-resolution cameras. The training set is randomly selected from 702 identities and the rest 702 pedestrians are utilized for testing. In addition, 408 additional dis-related identities are regarded as distractors.

\textbf{CUHK03-NP:} This dataset includes 14,097 images from 1,467 identities observed from 2 different cameras. There are two ways to obtain the annotations: manually labeled and DPM detected bounding boxes. For each camera, each person selects one image as the probe and we choose the rest images to construct the gallery set. The labelled dataset contains 767 identities, 7,368 training, 5,328 gallery and 1,400 query images while the detected set includes 767 identities, 7,365 training, 5,332 gallery and 1,400 query images.

\textbf{MSMT17:} It is a new person re-ID dataset, which includes 4,101 pedestrians and 126,441 bounding boxes. Different from other datasets, MSMT17 is randomly divided according to the ratio of training and testing 1:3. The training set includes 32,621 images with 1041 identities, while the testing set includes 93,820 images with 3,060 identities.

\begin{figure}[!t]
\centering
\includegraphics[width=3.15in]{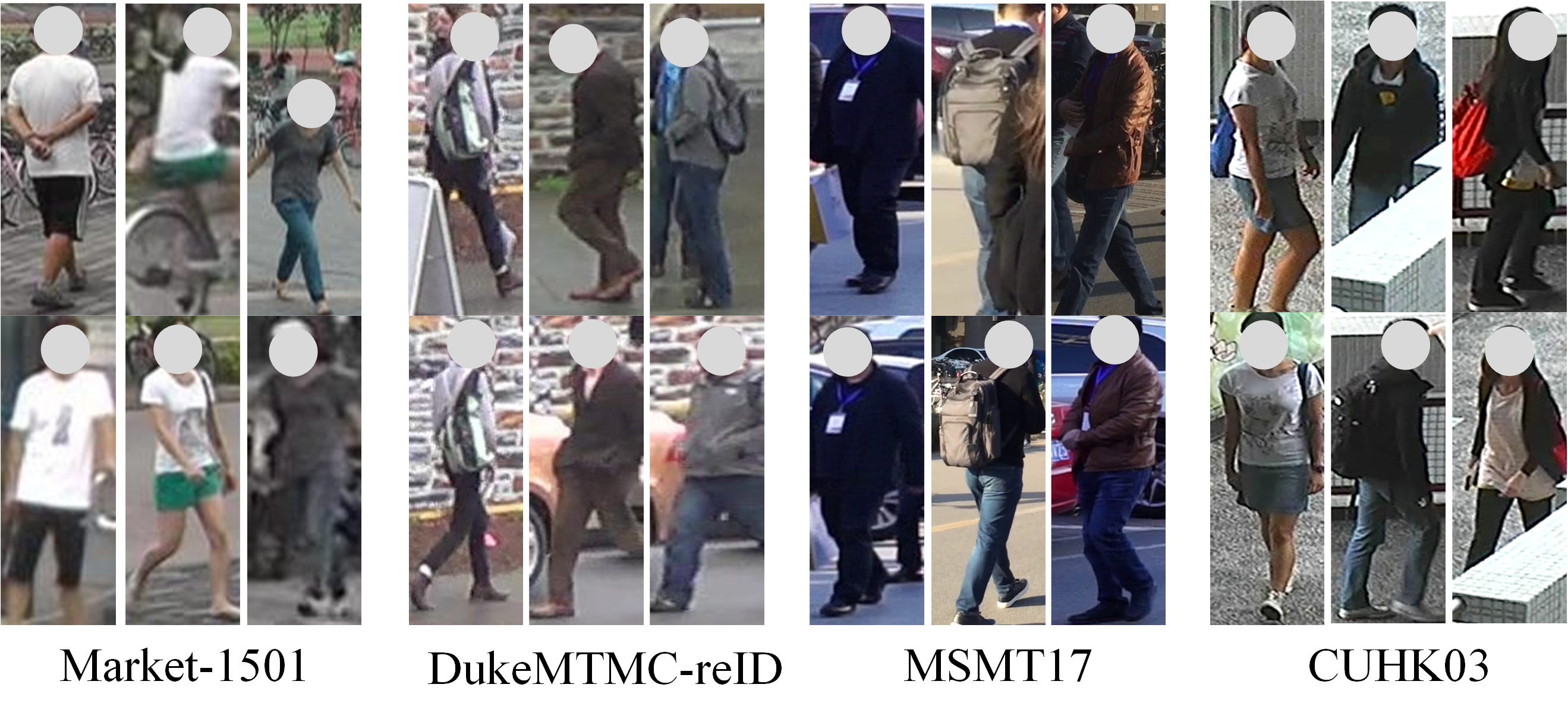}\\
\caption{Exemplary images in Market-1501, DukeMTMC-reID, MSMT17 and CUHK03 datasets. Each column represents the images of same person from different cameras.}
\label{fig:Images}
\end{figure}

Table~\ref{tab:1} lists four datasets widely adopted in person re-ID task and some images are shown in Figure~\ref{fig:Images}. All datasets contain many practical  challenges, such as occlusions, changes in viewpoint and lighting, or misaligned bounding boxes from object detectors.

\begin{table*}[!t]
\caption{All datasets are faced with some practical challenges: disappearing some parts of person due to occlusions, changes in light and viewpoint, or bounding box errors because of the object detectors. For the CUHK03-NP Dataset, L means manually labeled bounding boxes, D means DPM detected.}
\centering
\label{tab:1}
\begin{tabular}{l|c|c|c|c}
\hline
Dataset & Market & Duke & CUHK03-NP (L/D)  &MSMT17 \\
\hline
Identities   & 1501  & 1812   & 1467     &4101        \\
Bounding boxes & 32668 & 36411   & 13164   &126441      \\
Cameras   & 6     & 8      & 6         &   15             \\
Label method  & DMP/Hand  & Hand   & Hand/DPM   &Faster RCNN \\
Train images   & 12936  & 16522  & 7368/7365   &32621            \\
Train ids      & 751  & 702     & 767       &1041   \\
Test images    & 19732  & 17661  & 5328/5332   &93820 \\
Test ids       & 750   & 702    & 700    &3060     \\
\hline
\end{tabular}
\end{table*}

\subsection{Comparison with State-of-the-Art Approaches}
We compare our model with some advanced approaches on four datasets in this section.

\textbf{Market-1501:} It is a standard dataset for person re-ID. Bounding boxes of probe images in the dataset are manually drawn, while pedestrian bounding boxes in gallery are detected by using DPM detector. Table~\ref{tab:2} lists the results of our model and some advanced algorithms on Market-1501. From this table, we can find that our model obtains rank-1/mAP=95.7\%/87.7\% without using re-ranking algorithm. Table~\ref{tab:2} presents the current main approaches and all of which integrate global and local features in their networks. As can be seen from this table, our method obtains the best performance on rank-1 and competitive result on mAP. GCP obtains the highest accuracy on mAP and only 0.3\% improvement over our method.
\begin{table}[]
\caption{Comparisons (\%) on Market-1501 at 2 evaluation metrics: mAP, rank-1}
\label{tab:2}
\centering
\begin{tabular}{l|c|cc}
\hline
\multirow{2}{*}{Method} &\multirow{2}{*}{Backbone} &\multicolumn{2}{c}{Market-1501}          \\
\cline{3-4}
& & rank-1 & mAP       \\
\hline
DaRe (CVPR18)~\cite{34}  & ResNet50 & 86.4 & 69.3 \\
DaRe+RE (CVPR18)~\cite{34} & ResNet50  & 88.5 &74.2\\
PSE+ECN (CVPR18)~\cite{37}  & ResNet50  & 90.4   & 80.5 \\
HA-CNN (CVPR18)~\cite{26} & ResNet50 & 91.2  & 75.7     \\
DuATM (CVPR18)~\cite{38} &Inception-A & 91.4   & 76.6   \\
PCB+RPP (CVPR18)~\cite{39} & ResNet50 & 93.8  & 81.6     \\
MHN-PCB (ICCV19)~\cite{49} & ResNet50 &95.1 &85.0  \\
MGN (ACMMM18)~\cite{45} & ResNet50 &\textbf{95.7} & 86.9 \\
HPM (AAAI19)~\cite{71} & ResNet50 & 94.2 & 82.7 \\
AANet (CVPR19)~\cite{17} & ResNet152 &93.9 & 83.4 \\
DCDS(ICCV19)~\cite{50} & ResNet101 &94.8   & 85.8 \\
OSNet (ICCV19)~\cite{52}    &OSNeT &94.8   & 84.9   \\
GCP (AAAI20)~\cite{51}   & ResNet50 & 94.8  & \textbf{88.0}     \\
SAN(AAAI20)~\cite{53}    & ResNet50  & 95.1     & 85.8    \\
3DTANet (TCSVT20)~\cite{54}  &--   & 95.3   & 86.9     \\
HOReID (CVPR20)~\cite{56}   & ResNet50 &94.2 &84.9\\
RGA-CS(CVPR20)~\cite{57}  & ResNet50 &95.3  &87.8\\
\hline
\textbf{Ours}    & ResNet50   &\textbf{95.7}   &87.7  \\
\hline
\end{tabular}
\end{table}

\textbf{DukeMTMC-reID:} It is another standard dataset that contains a sufficient number of images for deep learning. Images in this dataset are of high quality and the pedestrian is complete. However, extra IDs are added to the dataset as a distraction for model training. The comparisons are reported in Table~\ref{tab:3}. Our model also performs very well on this dataset and achieves rank-1=90.2\%, mAP=80.2\% accuracy, which outperform state-of-the-art algorithms by a large margin. Our model achieves the best results on mAP and rank-1.

\begin{figure*}[!t]
  \centering
  \includegraphics[width=122mm]{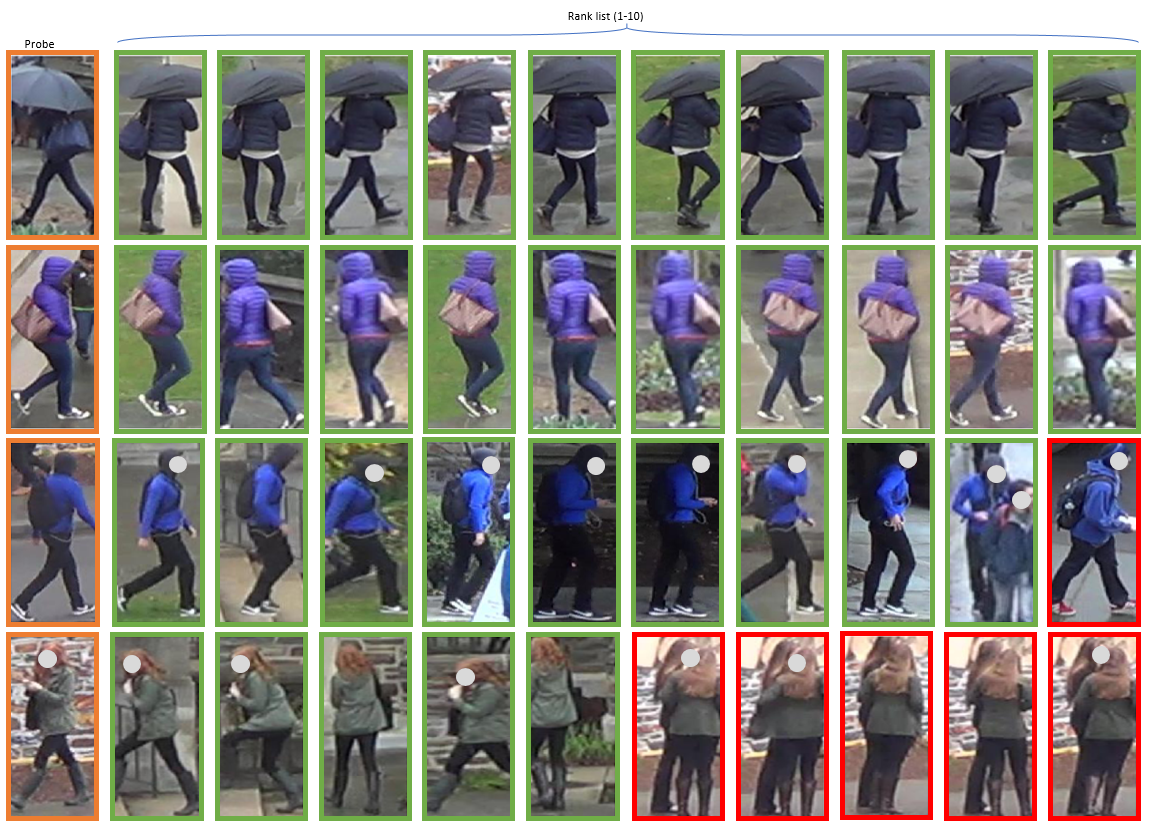}\\
  \caption{Visualization results on DukeMTMC-reID dataset. The first column is probe images and the right part lists top-10 retrieved gallery images corresponding to the probe image (not from the same camera). The images with green bounding boxes are the true matches, and those with red boxes are false ones.}
  \label{fig:ranklist}
\end{figure*}

The top-10 ranking results of some queries are displayed in Figure~\ref{fig:ranklist}. In this figure, the green bounding boxes list the correct match images and the red boxes list the wrong match images. The first three matching results show that our method has a strong robustness: the extracted features can identify the captured pedestrians well, regardless of their poses, views and lights changing. For the third query image, our method can still eliminate the interference of scene lighting changes to obtain accurate results. For the fourth query image, we can see that the current dataset contains a large number of similar images of pedestrians (ranks 1-10). Even with human eyes, it is impossible to distinguish whether these images belong to the same person, but our method can still obtain relatively accurate results (ranks 1-5).

\begin{table}[!t]
\caption{Comparisons (\%) on DukeMTMC-reid dataset at 2 evaluation metrics: mAP, rank -1}
\label{tab:3}
\centering
\begin{tabular}{l|c|ccc}
\hline
\multirow{2}{*}{Method} &\multirow{2}{*}{Backbone} &\multicolumn{2}{c}{DukeMTMC-ReID}          \\
\cline{3-4}
& & rank-1 & mAP       \\
\hline
SPReID (CVPR18)~\cite{58}    &ResNet152   & 85.9 & 73.3      \\
PCB+RPP (ECCV18)~\cite{39}  &ResNet50  & 83.3  & 69.2     \\
DuATM (CVPR18)~\cite{38} &DenseNet121   & 81.8   & 64.6       \\
PSE+ECN(CVPR18)~\cite{37} &ResNet50  & 84.5   & 75.7     \\
AANet (CVPR19)~\cite{17}  &ResNet152   & 87.7    &74.3     \\
DCDS(ICCV19)~\cite{50}    &ResNet101     & 87.6   & 75.5    \\
CASN(CVPR19)~\cite{60}    &ResNet50 &87.7 &73.7 \\
HPM (AAAI19)~\cite{71} & ResNet50 & 86.6 & 74.3 \\
MHN-PCB(ICCV19)~\cite{49}   &ResNet50 &89.1 &77.2\\
OSNet (ICCV19)~\cite{52}   &OSNET  &88.6 &73.5 \\
MGN(ACMMM18)~\cite{45} &ResNet50 &88.7 &78.4 \\
ABDNet (ICCV19)~\cite{61}  &ResNet50 &89.0 &78.6 \\
GCP(AAAI20)~\cite{51}  &ResNet50 &89.7  &78.6\\
SAN(AAAI20)~\cite{53}  &ResNet50 &87.9  &75.5\\
3DTANet (TCSVT20)~\cite{54}   &-- &89.9  &78.4 \\
M3+ ResNet50(CVPR20)~\cite{59}   &ResNet50 &84.7 &68.5 \\
M3+DenseNet121(CVPR20)~\cite{59}  &DenseNet121 &84.9 &68.0 \\
HOReID (CVPR20)~\cite{56}  &ResNet50 &86.9 &75.6 \\
\hline
\textbf{Ours}    &ResNet50   & \textbf{90.2}    & \textbf{80.2}    \\
\hline
\end{tabular}
\end{table}

\textbf{CUHK03-NP:} This is a challenging dataset. The challenge comes from the fact that it includes a large number of images with varied perspectives, pedestrian occlusions and low resolution which bring great interference to the training of the model. However, the proposed method still obtains the best results and surpasses all competing algorithms on rank-1. Our model also obtains the second best result on mAP. The comparisons are shown in Table~\ref{tab:4}. In our experiment, we adopt the new protocol of CUHK03 for training and testing. As can be seen when manually labeled bounding boxes are exploited, our method achieves rank-1=80.3\% and mAP=76.2\% accuracy. When detected setting are used, our method obtains rank-1=75.5\% and mAP=72.5\% accuracy.

\begin{table*}[]\small
\caption{Comparison results (\%) on CUHK03 dataset at 2 evaluation metrics: mAP, rank-1. L represents ‘labeled’ which means that the probe images are labeled by hand-crafted. D represents ‘detected’ which means that the probe images are labeled by DPM.}
\centering
\label{tab:4}
\begin{tabular}{l|c|cc|cc}
\hline
\multirow{2}{*}{Method} &\multirow{2}{*}{Backbone} & \multicolumn{2}{c|}{CUHK03 (L)} & \multicolumn{2}{c}{CUHK03 (D)}    \\
\cline{3-6}
&   & rank-1 & mAP  & rank-1 & mAP     \\
\hline
DaRe (CVPR18)~\cite{34} &DenseNet121   &56.4           &52.2    &54.3   & 50.1       \\
Mancs (ECCV18)~\cite{63}  &ResNet50  &69.0 &63.9 &65.5 &60.5\\
PCB+RPP(ECCV18)~\cite{39} &ResNet50 &-- &-- &63.7 &57.5\\
BFE(ICCV19)~\cite{62} &ResNet50 &79.4 &\textbf{76.7} &76.4 &\textbf{73.5}\\
MGN (ACMMM18)~\cite{45} &ResNet50   & 68.0  & 67.4  & 66.8   & {66.0}   \\
MHN-PCB(ICCV19)~\cite{49} &ResNet50 &77.2 &72.4 &71.7 &65.4\\
HPM (AAAI19)~\cite{71} & ResNet50 & -- & -- & 63.1& 57.5\\
OSNet (ICCV19)~\cite{52}  &OSNET  &72.3  &67.8 &-- &--\\
CASN(CVPR19)~\cite{60} &ResNet50 &73.7 &68.0 &71.5 &64.4 \\
M3+Res(CVPR20)~\cite{59} &ResNet50 &66.9 &60.7 &-- &--\\
M3+Dense(CVPR20)~\cite{59}  &DenseNet121 &61.6 &54.4 &-- &--\\
GCP(AAAI20)~\cite{51} &ResNet50 &77.9 &75.6 &77.9 &69.6\\
3DTANet(TCSVT20)~\cite{54} &--  &80.2 &75.2 &75.2 &68.9\\
\hline
Ours   &ResNet50  & \textbf{80.3}  & 76.2 & \textbf{77.5}   &72.5 \\
\hline
\end{tabular}
\end{table*}

\textbf{MSMT17:} This dataset is a new person re-ID dataset, which contains more pedestrians, more bounding boxes, and more cameras. It has more complex scenarios and backgrounds, such as outdoor and indoor. In addition, this dataset takes a long time to capture, covers multiple time periods, and has complex and obvious light changes. As a better pedestrian detector, Faster-RCNN was adopted to collect the dataset. Therefore, this dataset presents a more realistic situation, which is a great challenge for the current models. The comparison results with related methods are listed in Table~\ref{tab:5}. Our approach obtains the rank-1 of 79.6\% and 57.6\% on mAP, which is superior to most of related algorithms except ABDNet. Our model obtains the second best results on rank-1 and mAP.
\begin{table}[]
\caption{Comparisons (\%) on MSMT17 dataset at 3 evaluation metrics: mAP, rank-1 and rank-5.}
\label{tab:5}
\centering
\begin{tabular}{l|c|ccc}
\hline
\multirow{2}{*}{Method} &\multirow{2}{*}{Backbone} & \multicolumn{3}{c}{MSMT17}  \\
\cline{3-5}
&     & rank-1  &rank-5  & mAP        \\
\hline
GoogLeNet (ICCV17)~\cite{64} &GoogLeNet   &47.6 &-- &23.0     \\
PDC (ICCV17)~\cite{64} &GoogLeNet  &58.0  &73.6  &29.7   \\
GLAD (ACMMM17) ~\cite{66} &ResNet50   &61.4   &76.8  &34.0    \\
IANet (CVPR19) ~\cite{65} &ResNet50  & 75.5   & 85.5 & 46.8   \\
BFE (ICCV19) ~\cite{62} &ResNet50   & 78.8    &89.1   &51.5  \\
ABDNet (ICCV19)  ~\cite{61}   &ResNet50  & 82.3  &\textbf{90.6}  &\textbf{60.8}  \\
OSNet (ICCV19) ~\cite{52}  &OSNET  &78.7 &-- &52.9\\
SAN(AAAI20)  ~\cite{53} &ResNet50 &79.2  &-- &55.7\\
3DTANet (TCSVT20)  ~\cite{54} &-- &76.6  &86.8  &46.7\\
Circle Loss (CVPR20) ~\cite{67} &ResNet50 &76.3 &-- &50.2\\
\hline
\textbf{Ours}    &ResNet50   & \textbf{79.6}  &90.2  & 57.6  \\
\hline
\end{tabular}
\end{table}

\begin{table}[]
\caption{Ablation study on different components is evaluated on Market 1501 and DukeMTMC-reID datasets.}
\centering
\label{tab:6}
\begin{tabular}{l|cc|cc}
\hline
\multirow{2}{*}{Model} &\multicolumn{2}{c|}{Market-1501} & \multicolumn{2}{c}{DukeMTMC-reID}\\
\cline{2-5}  &rank-1  & mAP  & rank-1  & mAP\\
\hline
Baseline          & 88.1    & 71.1 &79.2  &63.7\\
+ Bag of Tricks   & 92.3    & 81.9 &85.1  &69.5\\
+ MPFE            & 93.4    & 83.4 &86.6  &73.5\\
+ MGO             & 94.7    & 86.5 &88.6  &77.9\\
+ SAM             & 95.3    & 87.2 &89.2  &78.1\\
+ CAM        & \textbf{95.7}   &\textbf{87.7} &\textbf{90.2}  &\textbf{80.2}\\
\hline
\end{tabular}
\end{table}

\subsection{Ablation Study}
To evaluate the usefulness of different components in our model, we perform several ablation experiments about each component in a single query mode on Market-1501 dataset. Note that all the experiments in this section follow the same settings in Section IV. A.

\textbf{Influences of Each Component (Same domain):} Our network is composed of different components, so we add these components (SAM, CAM, MPFE, MGO, and Bag of tricks) into the baseline in turn, and follow the same training settings \textcolor{black}{and bag of tricks consists of random erasing, Label Smoothing, and warm-up strategy.} The ablation experiment can prove that each component plays a positive role in improving the performance of our model.

The baseline we refer to in Section IV. A is ResNet-50, which can reach 71.1\% and 88.1\% of mAP and rank-1 on the Market-1501 dataset. The performance of our baseline used can reach the same level \textcolor{black}{compared} with the baseline reported by other papers. Then we add Bag of tricks, MGO, MPFE, SAM, and CAM to the baseline respectively during the training process. As shown in Table~\ref{tab:6}, the interaction of all these components makes our model reach 95.7\% on rank-1 and 87.7\% on mAP accuracy on Market-1501 dataset, and the individual experiment shows that each component has a positive effect \textcolor{black}{on} our model. Our approach has improved the baseline network by 7.6\% of rank-1 and 16.6\% of mAP. It is noteworthy that in all the components, the use of local features can greatly improve our model's performance, especially the improvement effect of mAP.

Meanwhile, we also introduce different components into the baseline one by one to evaluate how framework components contribute to the baseline. We conducted multiple experiments on the DukeMTMC-reID dataset. As shown in Table~\ref{tab:duke}, any single component can improve the performance of the model. This indicates that these five components are mutually complementary each other.
\textcolor{blue}{In addition, we keep the experimental settings unchanged and test the performance of MGO and PCB in our model, respectively. As can be seen from Table 7, when using PCB to obtain local features, the model obtains 88.7\% of rank-1 and 79.5\% of mAP. However, when using MGO to obtain the local visual cues, the model achieves 90.2\% of rank-1 and 80.2\% of mAP, and the results are improved by 1.5\% and 0.7\% for rank-1 and mAP, respectively. The experimental results show that the local features obtained by MGO are more robust and the feature representation ability of the model is also enhanced.}

\begin{table}[]
\caption{Ablation study on different components is evaluated on DukeMTMC-reID DATASET.}
\centering
\label{tab:duke}
\begin{tabular}{ccccccc|cc}
\hline
Baseline   & Tricks     & MGO      & MPFE       & SAM       & CAM       &\textcolor{blue}{PCB}    & rank-1 & mAP \\
\hline
\checkmark &            &           &           &           &            &          & 79.2   & 63.7\\
\checkmark &\checkmark  &           &           &           &            &          & 85.1   & 69.5\\
\checkmark &            &\checkmark &           &           &            &          & 84.9   & 69.5\\
\checkmark &            &           &\checkmark &           &            &          & 83.6   & 71.2\\
\checkmark &            &           &           & \checkmark&            &          & 84.3   & 72.3\\
\checkmark &            &           &           &           & \checkmark &          & 85.1   & 72.8\\
\textcolor{blue}{\checkmark} &\textcolor{blue}{\checkmark}  &\textcolor{blue}{\checkmark} &\textcolor{blue}{\checkmark} &\textcolor{blue}{\checkmark} & \textcolor{blue}{\checkmark} &          & \textcolor{blue}{90.2}   & \textcolor{blue}{80.2}\\
\textcolor{blue}{\checkmark} &\textcolor{blue}{\checkmark}  &  &\textcolor{blue}{\checkmark} &\textcolor{blue}{\checkmark} & \textcolor{blue}{\checkmark} & \textcolor{blue}{\checkmark}         & \textcolor{blue}{88.7}   & \textcolor{blue}{79.5}\\
\hline
\end{tabular}
\end{table}

\textbf{Influences of Each Component (Cross-domain):} To further verify the usefulness of each of our components, we also perform cross-domain experiments and the results are listed in Table~\ref{tab:7}. To eliminate the negative \textcolor{black}{effect} caused by overfitting, cross-domain experiments are conducted on Market-1501 and DukeMTMC-reID datasets. Duke $\rightarrow$ Market means training our model on DukeMTMC-reID dataset and evaluating it on Market-1501. Similarly, Market $\rightarrow$ Duke means that Market-1501 dataset is used for \textcolor{black}{training} and \textcolor{black}{DukeMTMC-reID} is used for evaluating. From the results, we can see that even in the cross-domain experiments, each of our components has \textcolor{black}{a} positive impact on the performance. Especially, on Duke $\rightarrow$ Market, Bag of tricks and MGO have huge performance improvements from rank-1/mAP=33.43\%/12.65\% to 41.21\%/17.13\% (+7.78\%/4.48\%) and from rank-1/mAP=42.31\%/18.90\% to 51.15\%/24.68\% (+8.84\%/5.78\%).
\begin{table}[!t]
\caption{Ablation study on different components is evaluated on cross-domain datasets. Duke $\rightarrow$ Market means trained on DukeMTMC-reID and evaluated it on Market-1501.}
\centering
\label{tab:7}
\begin{tabular}{l|cc|cc}
\hline
\multirow{2}{*}{Model} &\multicolumn{2}{c|}{Duke $\rightarrow$ Market} & \multicolumn{2}{c}{ Market $\rightarrow$ Duke}\\
\cline{2-5}  &rank-1  & mAP  & rank-1  & mAP\\
\hline
Baseline          & 33.43    & 12.65 &19.61  &9.19\\
+ Bag of \textcolor{black}{Tricks}   & 41.21    & 17.13 &30.48  &15.95\\
+ MPFE            & 42.31    & 18.90 &34.07  &18.22\\
+ MGO             & 51.15    & 24.68 &39.86  &23.01\\
+ SAM             & 51.98    & 25.16 &40.79  &23.96\\
+ CAM             & 54.33    & 26.43 &42.52  &25.37\\
\hline
\end{tabular}
\end{table}

\textbf{Effectiveness of Attention Module:} As shown in Table~\ref{tab:8}, Baseline$_{BMM}$ refers to \textcolor{black}{add} Bag of Tricks, MGO and MPFE modules to the baseline and aims to verify the effect of attention module on the performance of our model. We use S and C to respectively represent spatial attention module (SAM) and channel attention module (CAM) and '$\copyright$' represents connection operation. Therefore, S+C+$\copyright$ means we first adopt CAM and then use SAM, finally perform connection operation. The character order represents the order in which operations are performed. Similarly, ©+S+C means the first step is to perform connection operation. C+$\copyright$+S means the first step is to perform CAM. It can be seen from Table~\ref{tab:8} that Baseline$_{BMM}$+S+$\copyright$+C can get the best results and achieves 95.7\% on rank-1 and 87.7\% on mAP accuracy. The second part of this table shows the performance of our model after adding channel attention and spatial attention respectively (using the feature map of Stage 4 only). At the same time, we also consider the order of adding two attention modules into our model, which the results are shown in the third and fourth parts of this Table, respectively. We can see that when we first adopt spatial attention model and then perform connection, and finally adopt the channel attention model, our model can get the best results.

\begin{figure}[]
 \centering
 \includegraphics[width=3.15in]{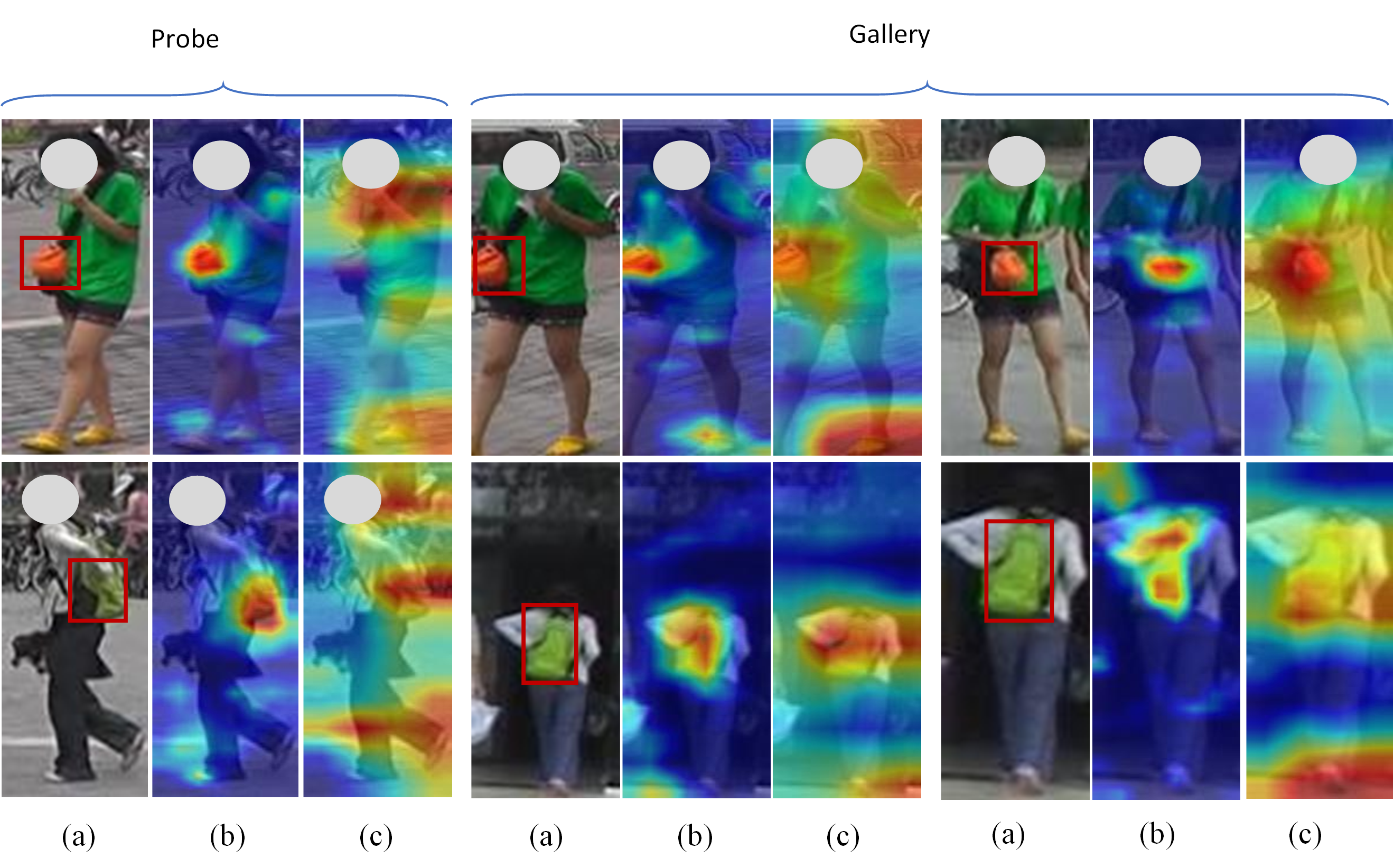}\\
 \caption{Each triplet contains from left to right, (a) original
image, (b) activation map of our model (Stage 4), and (c) activation map of baseline. These images show that our model is better able to focus on where is more discriminative feature.}
  \label{fig:7}
\end{figure}

Figure~\ref{fig:7} displays the activation maps of different person images \textcolor{black}{and we can see} that our model can focus on more discriminative \textcolor{black}{regions} than the baseline. \textcolor{black}{The first row of Figure~\ref{fig:7} shows that} our method can focus on the significant regions, regardless of changes of the perspective. In addition, our method can also detect the same significant \textcolor{black}{regions} with different perspectives and scales (the images in the second row).

\begin{table}[]
\caption{The performance of Attention Module is evaluated on Market1501 datasets.}
\centering
\label{tab:8}
\begin{tabular}{l|cc}
\hline
Model     & rank-1  & mAP  \\
\hline
Baseline$_{BMM}$    & 94.7    & 86.5 \\
\hline
Baseline$_{BMM}$ + S   & 94.9    & 86.6 \\
Baseline$_{BMM}$ + C   & 95.1    & 86.9 \\
\hline
Baseline$_{BMM}$ + $\copyright$+C+S  & 94.8    & 86.7 \\
Baseline$_{BMM}$ + C+S+$\copyright$   & 95.3   & 86.2 \\
Baseline$_{BMM}$ + C+$\copyright$+S   & 95.3    & 87.2 \\
\hline
Baseline$_{BMM}$ + $\copyright$+S+C  &95.0  &86.9 \\
Baseline$_{BMM}$ + S+C+$\copyright$  &95.5 &87.2 \\
Baseline$_{BMM}$ + S+$\copyright$+C  &\textbf{95.7} &\textbf{87.7}\\
\hline
\end{tabular}
\end{table}

\textbf{The impact of \textcolor{black}{Multi-resolution Feature} Fusion Strategy:} To evaluate the validity of the multi-\textcolor{black}{resolution feature} fusion, we divide our model into different stages and judge the validity of our model by fusing different stages of features. Table~\ref{tab:9} lists the results \textcolor{black}{of} our model at different stages, trained with random erasing and evaluated without re-ranking algorithm. By comparing the results of different stages, we can observe that as the stages gradually increase, the performance of our network gradually improves from rank-1/mAP= 84.80/63.67\% to 95.65/87.67\%. In addition, we also see that if we only consider single stage, with the \textcolor{black}{increased} number of stages, \textcolor{black}{a} deeper convolutional network can extract more discriminative features. However, note that stage 3 achieves higher performance than stage 4. The possible reason is that in stage 4, the features are too “high level” and too much information will be lost because of the extra pooling layers. As expected, the fusion of all the stages can get the best results, and mAP also gains a lot of improvements. This indicates that the robustness of the features after fusion has been greatly improved. What’s more, it is also worth noting that even if we only fuse a few stages of features such as stages 1-2, stages 1-3, and stages 2-4, our results still exceed most of related algorithms.

\begin{table}[!t]
\caption{Results (\%) with multi stages on Market-1501 and DukeMTMC-ReID datasets.}
\centering
\label{tab:9}
\begin{tabular}{l|cc|cc}
\hline
\multirow{2}{*}{Model} &\multicolumn{2}{c|}{Market-1501} & \multicolumn{2}{c}{DukeMTMC-reID}\\
\cline{2-5}  &rank-1  & mAP  & rank-1  & mAP\\
\hline
stage 1         & 84.80    & 63.67  &74.69  &55.67\\
stage 2         & 90.62    & 76.02  &82.50  &67.56\\
stage 3         & 93.89    & 83.43  &87.93  &75.36\\
stage 4         & 93.72    & 86.05  &88.63  &77.47\\
stage 1-2       & 90.80    & 76.06  &82.23  &68.00\\
stage 1-3       & 94.42    & 85.85  &88.33  &76.09\\
stage 2-4       & 95.24    & 86.93  &88.78  &78.01\\
\hline
All stage       & 95.65    & 87.67  &90.17  &80.16\\
\hline
\end{tabular}
\end{table}

\section{Conclusion}
In this paper, we \textcolor{black}{have} proposed a novel deeply supervised model for addressing the challenging person re-ID problem. By fusing the low- and high-level feature maps from any network, our model can effectively reduce the information loss. Our model can directly learn different granularity of local features from different stages, which are not used in some \textcolor{black}{parts} locating operations such as pose information. Hybrid-attention mechanisms are also introduced into our model to obtain more valuable features
at spatial and channel levels. Finally, We adopted two loss functions to train the network and learned the discriminative features for improving the matching performance. Experimental results display that our approach is superior to many state-of-the-art approaches.

\section*{Acknowledgment}
\textcolor{black}{This research is supported in part by the National Natural Science Foundation of China under Grant 61806099, U20B2065; and by the Natural Science Foundation of Jiangsu Province of China under Grant BK20180790; and by the Natural Science Research of Jiangsu Higher Education Institutions of China under Grant 18KJB520033; This research is also supported in part by the Priority Academic Program Development of Jiangsu Higher Education Institutions (PAPD) fund, in part by the Engineering Research Center of Digital Forensics, Ministry of Education.}

\end{document}